\documentclass[sigconf]{acmart}
\usepackage{arydshln}
\usepackage{multirow} 
\usepackage{booktabs}  
\usepackage{graphicx}
\usepackage{makecell}
\usepackage{xcolor}      
\usepackage{colortbl}    
\usepackage{amsmath}
\usepackage{threeparttable}
\usepackage{tabularx}
\usepackage[most]{tcolorbox}

\AtBeginDocument{%
  }

\renewcommand\footnotetextcopyrightpermission[1]{}

\newcommand{\METHODNAME}{\textsc{LPS-TC}}

\begin{document}

\title{Enabling Proactive Spoken Turns via a Generalized Style-Aware Full-Duplex Framework}


\author{
Tianrui Pan, Qinglin Zhang$^1$, Chong Deng$^1$, Luyao Cheng$^1$, Qian Chen$^1$, \\
Wen Wang$^1$, Jie Tang, Gangshan Wu, Jie Liu$^*$ \\
$^1$Token Foundry, Alibaba Group \\
}
\authornote{Corresponding author. liujie@nju.edu.cn}

\renewcommand{\shortauthors}{Trovato et al.}

\begin{abstract}

Compared with half-duplex dialogue systems where the system waits for user turn completion before it responds, natural full-duplex dialogue systems require agents to act proactively in real time, including timely interruptions and backchannels. This creates a key challenge: improving turn timing without sacrificing response quality. To address limitations in realistic proactive turn-taking, we build a generalized style-aware full-duplex framework with three key components. Firstly, we propose LPS-TC, a Lightweight Proactive Speech Turn Controller for plug-and-play integration. It features a fine-grained action space covering both reactive and proactive turn behaviors, enabling half-duplex models with full-duplex capabilities and enhancing existing full-duplex models with superior timing control. Secondly, we construct WildTurn, a large-scale, real-world English dataset containing approximately 2,981 hours of filtered multi-turn stereo conversations from face-to-face and telephone conversations, annotated with five turn-taking and five backchanneling styles. Trained on WildTurn, LPS-TC exhibits rich spoken dynamics that are not captured by existing static full-duplex benchmarks. Thirdly, we introduce a two-tier evaluation scheme that assesses both chunk-level timing precision and turn-level interaction quality under realistic streaming constraints. Our experiments, integrating LPS-TC with half-duplex models like Qwen2.5-Omni and full-duplex models like Freeze-Omni, showcase its superior performance in timing appropriateness and response quality. Our framework also demonstrates fine-grained style controllability and strong generalizability, enabling more natural and human-like spoken interactions.

\end{abstract}



\maketitle

\section{Introduction}

\begin{figure}
    \centering
    \includegraphics[width=1.0\linewidth]{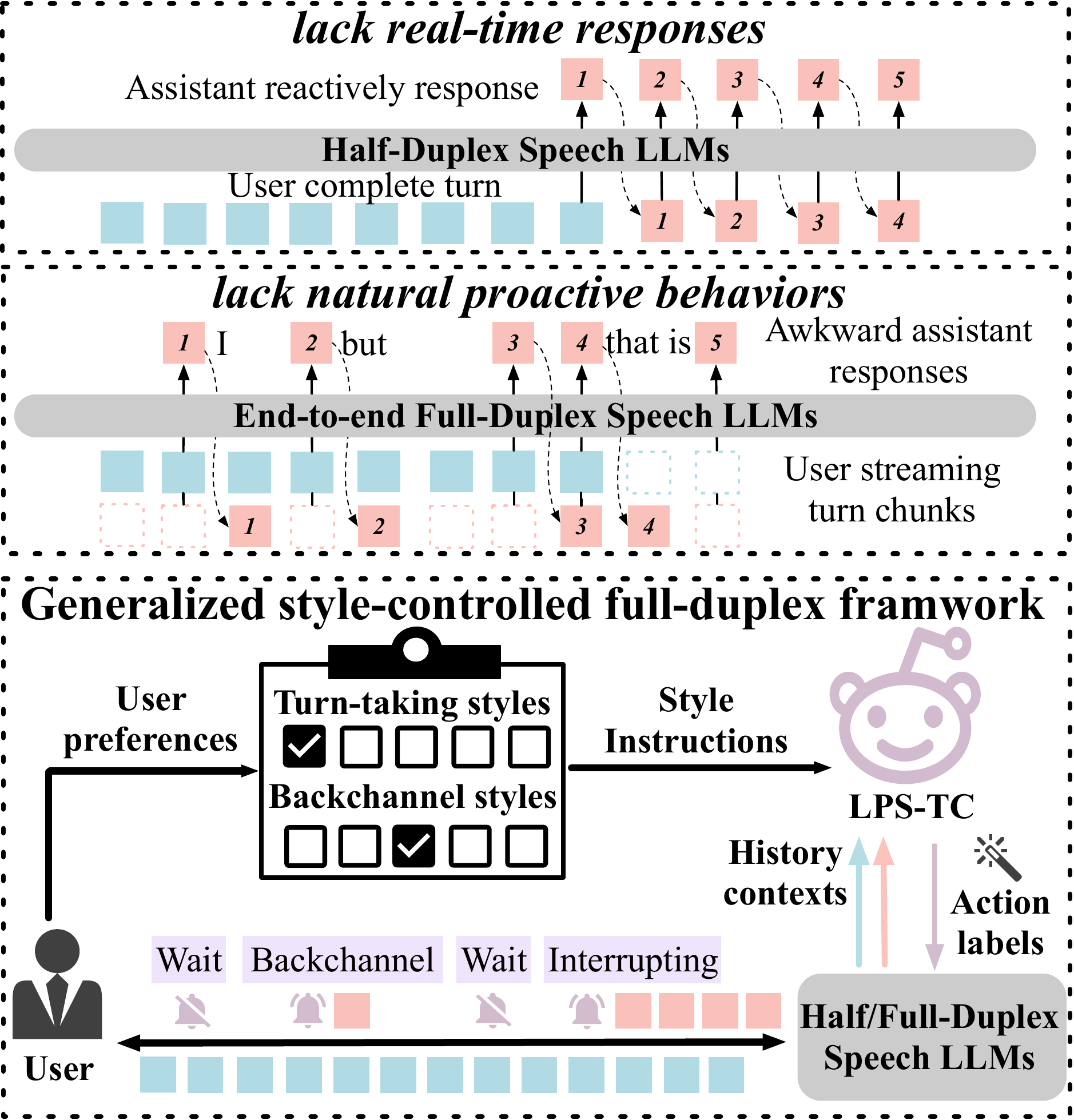}
    \caption{Proposed generalized full-duplex dialogue framework. Our lightweight, plug-and-play turn controller LPS-TC (i) integrates seamlessly with off-the-shelf half/full-duplex Speech LLMs, (ii) supports style-conditioned proactive spoken behaviors such as backchannels and interruptions.}
    \label{fig:introduction}
\end{figure}

Spoken dialogue systems have progressed from text-based dialogue frameworks that emphasize contextual understanding~\cite{chen2022unidu,liao2021dialogue, wu2020tod} and response generation~\cite{roller2021recipes,ye2022reflecting} to timing-sensitive, expression-rich conversational agents. While some works develop turn-based half-duplex voice assistants~\cite{nguyen2023generative,wu2025step} into full-duplex models that can listen and speak concurrently~\cite{wang2024freeze,wangntpp}, other lines of work investigate diverse spoken behavioral styles, including distinct personalities and fine-grained emotional expression~\cite{tu2025ultravoice, geng2025osum, cui2025recent}. Different from reactively following the user-oriented conversation, natural full-duplex systems should manage real-time proactive turn behaviors such as interrupting for clarifications, offering information without being asked, or backchannels to show engagement. Such behaviors enable assistants to support humans not only mechanically but also socially and emotionally. However, research on proactivity in spoken dialogues remains underexplored. Some works~\cite{deng2025proactive, dengplug} consider the semantics of text-based responses, while others~\cite{nguyen2023generative, mitsui2023towards, Behavior-SD} struggle with heterogeneous user preferences, since the same proactive responses may be perceived as supportive by some users yet intrusive by others. Enabling strategic and motivational turns in natural spoken dialogues requires fine-grained temporal grounding that jointly accounts for semantics, prosody, and interaction style, yet existing spoken dialogue models still lack this capability~\cite{chang2025game}. Yet existing efforts remain fragmented across turn-control methods, datasets, and evaluation protocols, leaving no unified support for proactive, style-controllable full-duplex interaction. To bridge this gap, we propose a generalized full-duplex framework with three key components.

\textbf{First, we propose \METHODNAME{}, balancing the trade-off between high response quality and natural spoken turn interactions with its architectual design.} While advanced half-duplex speech large language models (LLMs)~\cite{wu2025step,yang2025qwen3} excel in response quality, their strictly sequential turn-taking limits real-time processing and natural spoken-turn dynamics. Meanwhile, developing effective full-duplex models remains challenging. Some proprietary commercial agents~\cite{openai2024gpt4ocard, intelligence2025amazon} incur prohibitive costs for precise timing, while open-source solutions~\cite{wang2024freeze, chen2025fireredchat} suffer from limited interactional modeling and degradation in response quality. Our \METHODNAME{}, depicted in Figure~\ref{fig:introduction}, balances this trade-off. It is a lightweight, plug-and-play spoken turn controller for turn-timing prediction and style-conditioned spoken interactions. It autoregressively processes dual-channel audio streams from both user and assistant. This allows it to capture critical paralinguistic cues and conversational dynamics, which are typically lost in text-centric or single-stream models. \METHODNAME{} can either equip half-duplex speech LLMs with full-duplex capability or enhance existing full-duplex models with superior timing control.

\textbf{Second, we introduce WildTurn, an English dataset with fine-grained proactive spoken turn dynamics.}  Most available spoken dialogue datasets~\cite{lee2023dailytalk, lin2024advancing, tu2025ultravoice, geng2025osum, cui2025recent} mainly focus on utterance-level behavioral or paralinguistic styles. They rarely capture interactional dynamics such as overlapping speech. Although Behavior-SD~\cite{Behavior-SD} incorporates turn behaviors, it is synthetic and may not capture the diversity of real-world conversations. To address these limitations, we curate WildTurn from \textbf{real-world} interactions and annotate it with a comprehensive label space encompassing five action categories: Normal Turn Taking (NTT), Interruptive Turn Taking (ITT), Backchanneling (BC), Barge In (BI), and No Action (NA).  Among these, NTT and BC are further refined into style labels: we define five turn-taking styles for NTT and five backchannel styles for BC based on statistical metrics such as silence or overlap duration and action frequency. We then use GPT-5.2~\cite{singh2025openai} to generate style-conditioned instructions. Moreover, to capture timing in human-human conversations, we employ span-based labeling informed by human-computer interaction studies~\cite{hci_1,hci_2} on realistic intent-to-speech latency, with each label spanning from cue to response onset.

\textbf{Third, we address the limitations of current full-duplex benchmarks and build a new evaluation protocol towards more holistic and real-time interactions.} Prior full-duplex benchmarks~\cite{lin2025full, peng2025fd, arora2025talking} often overlook two aspects. First, they exclude advanced half-duplex speech LLMs such as Qwen3-Omni~\cite{yang2025qwen3}: when adapted with proactive prompts, these models can serve as powerful proactive controllers for a fair comparison. Second, their reliance on static, pre-recorded dialogues creates a fundamental \textit{context mismatch}: in multi-turn settings, each subsequent user turn is influenced by the previous ground-truth assistant response in both timing and content. To address these issues, we propose a two-tier framework that evaluates \textbf{chunk-level timing precision} and \textbf{turn-level interaction quality} across turn-taking, backchanneling, and turn-yielding under \textbf{realistic streaming constraints}. We benchmark against a broad spectrum of half-duplex baselines via standardized proactive prompting. To resolve the context mismatch, we reconstruct the test set by splitting multi-round dialogues into individual turns. Each turn is then presented to the model along with its real preceding context. Experimental results show that \METHODNAME{} outperforms other turn controllers in \textbf{chunk-level} timing accuracy based on our labeling space. It also improves \textbf{turn-level} interaction quality when integrated with both half-duplex and full-duplex speech LLMs. In addition, \METHODNAME{} demonstrates superior style controllability and robust instruction following. 


\section{Related Work}
\textbf{Full-Duplex Spoken Turn Controller.} End-to-end full-duplex models~\cite{defossez2024moshi, zhang2025omniflatten, wangntpp, yu2024salmonn} can handle overlapping speech, but their tightly coupled design is costly to train and may degrade response quality~\cite{xie2024mini,chen2025reinforcement,arora2025chain,roy2026personaplex}. To address this issue, other works add a turn controller, either VAD-based~\cite{wang2024turn, fu2025vita, mai2025real, wang2024freeze} for binary state modeling or hidden-state-based~\cite{chang2022turn, ma2025language, chen2025minmo, liu2025x, lu2025cleans2s} for richer turn prediction. Compared with standard VAD methods~\cite{Silero-VAD, wu2025phoenix, xu2026fireredasr2s}, which only model binary turn states, recent controllers~\cite{semanticvad, li2025easy, liao2025flexduo} add states such as \textit{wait} and \textit{idle} to better handle backchannels and background noise. In contrast, \METHODNAME{} decouples timing control from response generation, enabling fine-grained assistant behaviors such as interruptions and backchannels without sacrificing response quality.

\textbf{Proactive Spoken Interactions.} Although proactivity in conversational agents has attracted growing attention~\cite{zargham2022understanding, deng2023survey, liao2023proactive, zhang2024proagent}, most prior work focuses on text, while spoken floor-taking remains underexplored~\cite{viswanath2026desirability}. Existing efforts study either real-time agent frameworks~\cite{qiu2026building, livekit_agents_2024, pipecat_2024, huggingface_s2s_2024} or sentence-level speaking styles~\cite{tu2025ultravoice, geng2025osum, cui2025recent}, but conversational fluency also depends on \textit{diverse spoken turn behaviors}. We therefore focus on turn-level proactive spoken interactions, including timely backchannels, predictive turn-taking, and smooth turn-yielding. However, existing spoken dialogue datasets with proactive behaviors~\cite{nguyen2023generative, mitsui2023towards, Behavior-SD, zhou2025open} remain limited. They rely on synthetic data, which fail to capture the nuanced dynamics of real conversations~\cite{Behavior-SD}. We construct and annotate WildTurn from large-scale real-world face-to-face and telephone conversations with diverse turn-taking and backchannel styles.

\textbf{Full-duplex benchmarks.} Existing full-duplex benchmarks mainly focus on isolated or limited spoken interactions. Early work such as Full-Duplex-Bench~\cite{lin2025full,lin2025full1v5} evaluates pause handling, interruption response, and overlap management, but is largely restricted to single-round settings. Later benchmarks extend to multi-round scenarios, yet still lack two aspects: evaluating generalized Speech LLMs as real-time turn-taking models under streaming input, and modeling multi-turn dynamic interaction under realistic streaming conditions. For example, Talking Turns~\cite{arora2025talking} focuses on timing prediction, FD-Bench~\cite{peng2025fd} emphasizes interruption-heavy cases, and Full-Duplex-Bench-v2~\cite{lin2025fullv2} relies on a separate Speech LLM as examiner. In contrast, we propose a two-tier real-time evaluation framework that assesses generalized Speech LLMs at both chunk and turn levels in realistic multi-turn streaming interactions.

\begin{figure*}
    \centering
    \includegraphics[width=1.0\linewidth]{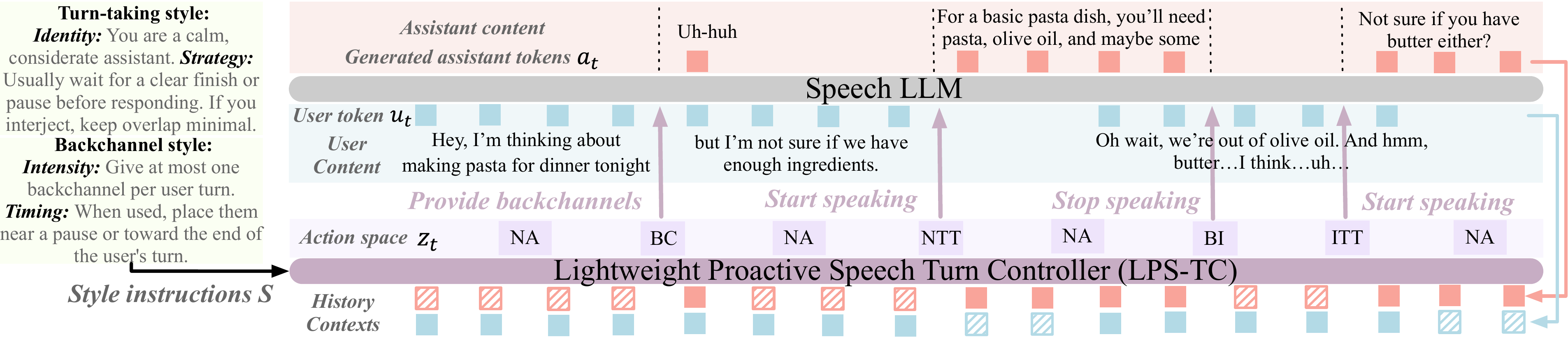}
    \caption{Decoupled full-duplex pipeline. LPS-TC predicts action $z_t$ based on style instruction $S$ and history contexts of both user input $U_{a,<t}$ and SLM-generated audio $A_{a,<t}$. SLM generates $a_t$ from $U_{a,<t}$ and $z_t$. LPS-TC offers an expanded action space with both reactive and proactive behaviors, facilitating precisely-timed backchannels and interruptions.}
    \label{fig:decoupled_full_duplex_pipeline}
\end{figure*}

\section{Method}

We present a unified full-duplex framework for natural proactive spoken interaction with style-aware, fine-grained turn control. Section~\ref{sec:problem_formalization} formalize the differences between half-duplex and full-duplex interaction, Section~\ref{sec:model_pipeline} introduces \METHODNAME{} for turn control, and Section~\ref{sec:style-label} presents WildTurn data construction pipeline.

\subsection{Problem Formalization}
\label{sec:problem_formalization}

Reactive half-duplex and proactive full-duplex systems mainly differ in the history context $H$ they condition on. We discretize the audio stream into small chunks indexed by time step $t$. A \textbf{half-duplex model} conditions on the complete user utterance as a fixed context. Let $U_a = \{u_0, \dots, u_{T-1}\}$ denote the full user audio utterance. The model conditions on a history context $H_{\text{half}}$ that is available only after the user finishes speaking, i.e., at time $T$. The assistant response generation of $A_a = \{a_1, \dots, a_M\}$ is thus formalized as:
\begin{equation}
    P(A_a | H_{\text{half}}) = \prod_{m=1}^{M} P(a_m | a_{<m}, H_{\text{half}}),\space \space H_{\text{half}} = \text{Context}(U_a).
    \label{eq:half-duplex}
\end{equation}
where $\text{Context}(\cdot)$ is a function mapping the available interaction history to a contextual representation. This formulation is inherently reactive: the fixed context $H_{\text{half}}$ cannot capture real-time dynamics and thus cannot support interruption or overlapping speech. A \textbf{full-duplex model} operates on a dynamic context $H_{\text{full},t}$ that evolves at each time step $t$, comprising all available information: the incoming user audio $U_{a, < t}$ and the model's own audio history $A_{a, <t}$.
\begin{equation}
    H_{\text{full}, t} = \text{Context}(U_{a, < t}, A_{a, <t}).
\end{equation}
At each time step, the model implicitly decides whether to speak or wait. This decision can be formalized as a joint probability over a conceptual timing variable $z_t \in \{\text{SPEAK}, \text{SILENCE}\}$ and the output audio token $a_t$:
\begin{equation}
    P(z_t, a_t | H_{\text{full}, t}) = P(z_t | H_{\text{full}, t}) \cdot P(a_t | z_t, H_{\text{full}, t}).
    \label{eq:full-duplex}
\end{equation}
In a fully integrated model, the conceptual variable $z_t$ is not an explicit architectural component; instead, it implicitly guides the output generation. Eq.~\ref{eq:full-duplex} thus reveals a core challenge: an integrated model must simultaneously handle two interconnected tasks: \textbf{timing control} (determining $z_t$) and \textbf{content generation} (predicting $a_t$). This joint optimization causes a trade-off, where improving low-latency prediction of $z_t$ will compromise the complex reasoning required for high-quality generation of $a_t$, and vice versa.

\subsection{Style-Aware Decoupled Full-Duplex Pipeline}
\label{sec:model_pipeline}
To handle the trade-off between timing $z_t$ and response quality $a_t$, we introduce \textbf{\METHODNAME{}}, a Lightweight plug-and-play Proactive Speech Turn Controller. It factorizes the joint decision in Equation~\ref{eq:full-duplex} by explicitly modeling the timing variable $z_t$ separately. Prior turn controllers~\cite{yu2024salmonn,liao2025flexduo} either serialize overlap into an interleaved user-assistant token sequence or feed only user speech to the model. These designs increase latency and fail to preserve paralinguistic cues that are critical for bidirectional interaction. Moreover, because user preferences are heterogeneous, the same proactive action may be perceived as welcome by some users but unwelcome by others. \METHODNAME{} introduces two key innovations:
\begin{itemize}
    \item Low-latency dual-channel architecture, which directly processes raw audio from both user ($U_a$) and assistant ($A_a$) for simultaneous interaction without additional preprocessing.
    \item Style-aware turn control, which incorporates an explicit style instruction $S$ to adapt turn-taking and backchannel behavior to user preferences.
\end{itemize}
For streaming inference, we adapt the Whisper encoder with causal convolutions~\cite{van2016wavenet} and block causal attention~\cite{zeng2024glm}. In Figure~\ref{fig:decoupled_full_duplex_pipeline}, at each time step $t$, \METHODNAME{} (denoted by $f$) explicitly predicts $z_t$ conditioned on the accumulated context and the given style:
\begin{equation}
    \label{eq:LPS-TC_formula}
    z_t = f(U_{a,< t}, A_{a,< t},  Z_{<t}, S),
\end{equation}
where $z_t \in \{\text{NA, NTT, ITT, BC, BI}\}$ represents the predicted action: No Action (\textbf{NA}), Normal Turn Taking (\textbf{NTT}), Interruptive Turn Taking (\textbf{ITT}), Backchannel (\textbf{BC}), or Barge In (\textbf{BI}). The style instructions $S$ consist of two components: the turn-taking style $S_{\text{TT}}$ controls the turn-taking tendency across five levels, and the backchannel style $S_{\text{BC}}$ specifies one of five feedback patterns based on frequency and timing. Details are given in Section~\ref{sec:style-label}. The comprehensive action space and style conditioning together allow \METHODNAME{} to provide nuanced and adaptive turn control. Then the SpeechLLM, denoted by $F$, predicts response speech tokens $a_t$ as follows:
\begin{equation}
    a_t = F\bigl(z_t, H_{\text{half},t}|H_{\text{full},t}),
\end{equation}
where the predicted action $z_t$ from \METHODNAME{} determines the behavior of SpeechLLM. Specifically, NTT, ITT, and BC trigger SpeechLLM to generate speech responses, NA instructs SpeechLLM to maintain its current state (i.e., either continue speaking or remain silent), and BI instructs SpeechLLM to stop generating speech tokens. After SpeechLLM executes the action at time $t$, the current user input $u_t$ and assistant output $a_t$ are appended to the history (forming $A_{a,t}$ in Equation~\ref{eq:LPS-TC_formula}), which then predicts the next action $z_{t+1}$. In this way, \METHODNAME{} and SpeechLLM form a closed real-time control loop. Theoretically, \METHODNAME{} can equip half-duplex models with full-duplex capability by expanding their context $H_{\text{half}}$ (Eq.~\ref{eq:half-duplex}) with the assistant's audio responses. It can also enhance full-duplex models by replacing the implicit timing variable $z_t$ (Eq.~\ref{eq:full-duplex}) with a broader action space that includes ITT and BC in addition to NTT, BI and NA. In summary, the decoupled and style-aware design of \METHODNAME{} enables low-latency, fine-grained turn control while supporting adaptive, personalized spoken interactions.

\begin{table*}[!h]
\setlength{\tabcolsep}{3pt}
\centering
\small
\caption{Style definitions for five turn-taking styles $S_{\text{TT}}$ and five backchannel styles $S_{\text{BC}}$, based on statistical thresholds for turn-boundary delays and action frequencies, which are derived from the spoken turn action labels (NA, NTT, ITT, BC, BI).}

\label{tab:style_prompts}
\begin{tabularx}{\textwidth}{@{}llX@{\hspace{6pt}}lll@{}}

\toprule
\multicolumn{3}{c}{\textbf{Turn-taking}} & \multicolumn{3}{c}{\textbf{Backchannel}} \\
\cmidrule(r){1-3} \cmidrule(l){4-6}
\textbf{Style} & \textbf{Ratio} & \textbf{Boundary timing} & \textbf{Style} & \textbf{Frequency} & \textbf{Onset Timing} \\
\midrule
Patient & Only NTT actions. & N/A & High-Early & High BCs/turn or high BCs/minute & Shortly after user starts \\
\cmidrule(r){1-3} \cmidrule(l){4-6}

$\text{Mixed}_{\text{low}}$ & Both ITT and NTT. & Higher NTT latency, shorter ITT lead. & High-Late & High BCs/turn or high BCs/minute & Near pause or user end \\
\cmidrule(r){1-3} \cmidrule(l){4-6}

$\text{Mixed}_{\text{medium}}$ & Both ITT and NTT. & neutral NTT latency and ITT lead time. & Low-Early & Low BCs/turn and low BCs/minute & Shortly after user starts \\
\cmidrule(r){1-3} \cmidrule(l){4-6}

$\text{Mixed}_{\text{high}}$ & Both ITT and NTT. & Lower NTT latency, longer ITT lead. & Low-Late & Low BCs/turn and low BCs/minute & Near pause or user end \\
\cmidrule(r){1-3} \cmidrule(l){4-6}

Assertive & Only ITT actions. & N/A & No Backchannel & \multicolumn{2}{c}{No BC actions.} \\

\bottomrule
\end{tabularx}
\end{table*}

\subsection{WildTurn Dataset Construction}
\label{sec:style-label}

\begin{figure}
    \centering
    \includegraphics[width=1.0\linewidth]{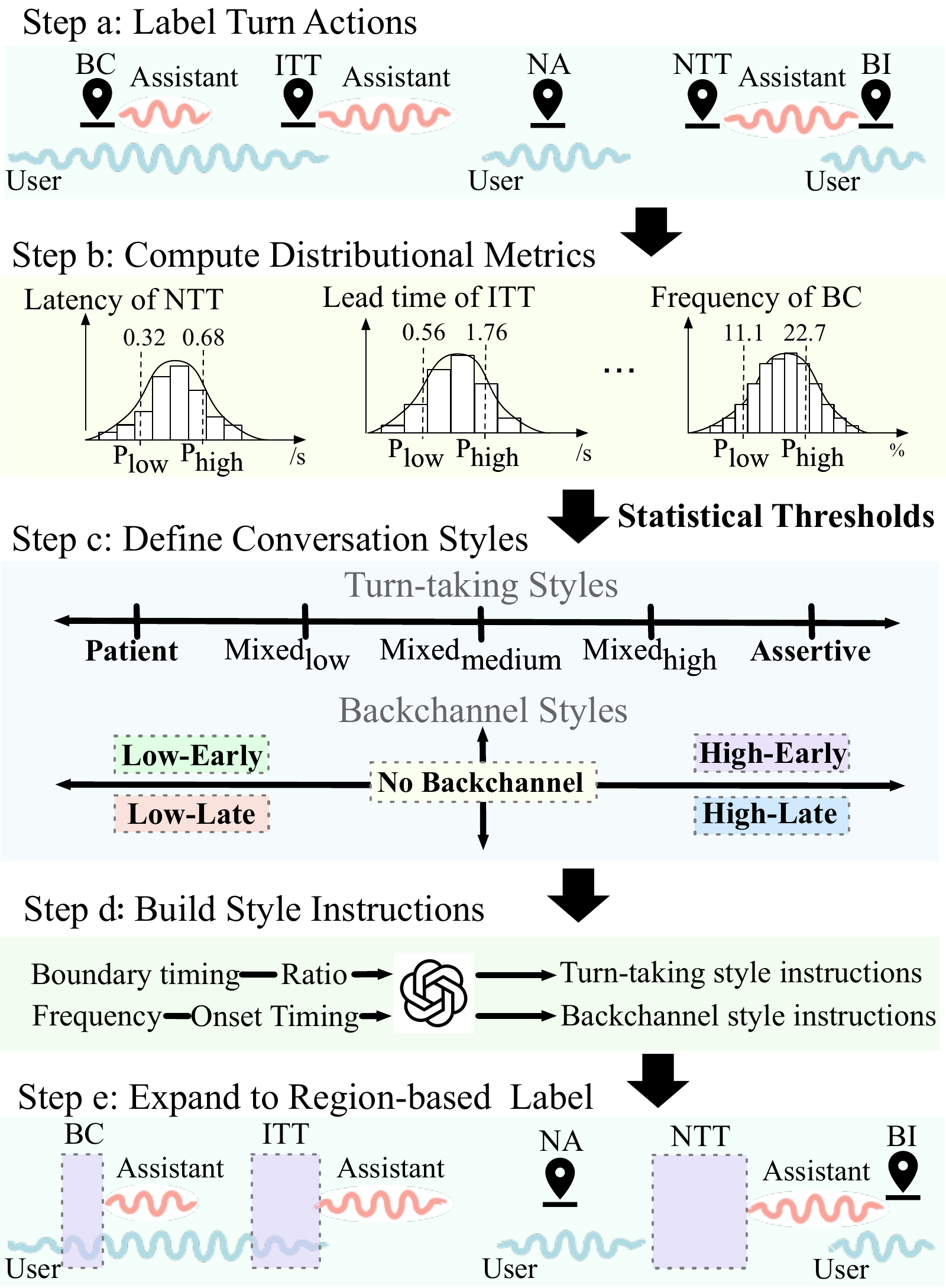}
     \caption{Overview of the WildTurn construction pipeline: chunk-level action labeling, metric extraction, style categorization, instruction generation, and region-based expansion.}
    \label{fig:dataset_pipeline}
\end{figure}

Existing dialogue datasets are insufficient for learning natural proactive spoken turns. Most focus on utterance-level expressive styles, lack fine-grained reactive and proactive turn labels, and often rely on synthetic data~\cite{lee2023dailytalk, lin2024advancing, Behavior-SD}. We construct \textbf{WildTurn}, a real-world dataset with multi-turn conversations, fine-grained spoken turn action labels $Z$, and corresponding style instructions $S$. Figure~\ref{fig:dataset_pipeline} summarizes the five-step construction pipeline.

\textbf{Step a: Label spoken turn actions.} We assign fine-grained action labels at the chunk level. Following~\cite{arora2025talking}, each 40\,ms audio chunk is annotated with one action label, matching the temporal resolution of the tokenized audio input. We first identify assistant state transitions between \textsc{Speak} and \textsc{Silence}. We then assign NTT, ITT, BC, or BI at each transition boundary, while all remaining chunks are labeled as NA (Figure~\ref{fig:dataset_pipeline}a). NTT, ITT, and BI are detected directly from VAD-based~\cite{Silero-VAD} state changes, whereas BC is identified with a lexicon-based procedure using a 66-entry English backchannel lexicon expanded from~\cite{ekstedt22_interspeech}. On average, each dual-channel audio sample in WildTurn contains 2.26 interruptions (ITT), 1.70 backchannels (BC), and 3.64 natural turn-takes (NTT).

\textbf{Step b: Compute distributional metrics.} Based on the action labels from Step a, we define a small set of conversation-level behavioral metrics. For turn-taking, these include the ITT-to-NTT ratio, NTT latency, and ITT lead time. For backchanneling, they include backchannel frequency and onset timing. Figure~\ref{fig:dataset_pipeline}b shows the corresponding thresholds obtained by quantile-based partitioning over their empirical distributions.

\textbf{Step c: Define conversation styles.} Using the metrics from Step b, we map each conversation to discrete style categories for turn-taking and backchanneling, as shown in Figure~\ref{fig:dataset_pipeline}c. For \textbf{turn-taking styles}, the ratio and boundary timing metrics define a spectrum from fully \textit{Patient} behavior to fully \textit{Assertive} behavior. We further define three intermediate categories: \textit{$\text{Mixed}_{\text{low}}$}, \textit{$\text{Mixed}_{\text{medium}}$}, and \textit{$\text{Mixed}_{\text{high}}$}. For \textbf{backchanneling styles}, backchannel frequency and onset timing form a two-dimensional grid. This yields four active styles, namely \textit{High-Early}, \textit{High-Late}, \textit{Low-Early}, and \textit{Low-Late}, together with the \textit{No Backchannel} style. Table~\ref{tab:style_prompts} summarizes the resulting style definitions.

\textbf{Step d: Build style instructions.} Based on the style categories defined in Step c, we use GPT-5.2~\cite{singh2025openai} to verbalize the metric-based style definitions in Table~\ref{tab:style_prompts} into concise natural-language instructions for model conditioning. These instructions serve as the style input paired with the action labels in WildTurn.

\textbf{Step e: Expand to region-based labels.} Finally, we convert the point-wise action labels from Step a into region-based labels for temporally tolerant supervision. Since point-wise labels are sparse and sensitive to onset errors, we extend each action label backward from its onset to form an active region, following prior human-computer interaction studies on the delay between speaking intention and speech onset~\cite{hci_1,hci_2} (Figure~\ref{fig:dataset_pipeline}e).

\section{Experiments}
\subsection{Datasets and Implementation Details}
WildTurn is built from 1.2k hours of face-to-face conversations from Seamless Interaction~\cite{agrawal2025seamless}, 2k hours of telephonic conversations from Fisher~\cite{cieri2004fisher}, and a small amount of synthetic data from Behavior-SD~\cite{Behavior-SD} for training stability. The dataset totals 2,981 hours of filtered stereo audio and 86,430 samples. Original recordings are segmented into clips of up to 120 seconds using GPT-4o~\cite{openai2024gpt4ocard} to identify natural breakpoints such as topic shifts or speaker restarts. We split the data into training, validation, and test sets with 85323, 100, and 1007 samples, respectively. \METHODNAME{} comprises an audio encoder initialized from Whisper-large-v3~\cite{radford2023robust}, an audio adapter, and a Qwen3-0.6B~\cite{yang2025qwen3} backbone. For efficiency, we use a 20-second sliding window over the most recent audio context. 

\subsection{Two-tier Real-time Full-Duplex Evaluation}
\label{sec:eval}

We propose a \textbf{two-tier} real-time full-duplex evaluation scheme that measures \textbf{chunk-level} timing precision and \textbf{turn-level} interaction quality for turn-taking, backchanneling, and turn-yielding. Compared with prior full-duplex benchmarks, our evaluation scheme has two key differences: 
\begin{itemize}
    \item We treat generalized Speech LLMs as real-time turn-taking models by evaluating incremental chunk-level action prediction over streaming audio.
    \item We segment the test set into streaming-aligned evaluation instances, each paired with its preceding dialogue history. This enables both chunk-level and turn-level evaluation under realistic streaming conditions.
\end{itemize}

\textbf{Assistant Turn-taking Evaluation.}
We formulate turn timing prediction for generalized Speech LLMs as incremental action prediction over audio chunks. For each \textbf{user} turn $u_t$, given conversation history $H_{t-1} = \{(u_1, a_1), \ldots, (u_{t-1}, a_{t-1})\}$ and streaming audio chunks $C_t = \{c_1, \ldots, c_n\}$, where each chunk $c_i$ is a 640\,ms segment chosen to balance temporal resolution and model robustness, the model predicts an action at each step $i$:
\begin{equation}
    \label{eq:slm_as_turn_timing_prediction}
    \hat{z}_i = F(H_{t-1}, c_{1:i}),
\end{equation}
where $\hat{z}_i \in \{\textit{wait}, \textit{backchannel}, \textit{response}\}$. These actions are mapped to the interaction labels in Equation~\ref{eq:LPS-TC_formula}: \textit{wait} to NA, \textit{backchannel} to BC, and \textit{response} to either ITT or NTT. A \textit{response} prediction is categorized as ITT if its first onset $i^*$ occurs before the end of the user turn (i.e., $i^* < n$), and as NTT otherwise. Here, $i^* = \min \{i \mid \hat{z}_i = \textit{response}\}$ denotes the index of the first response. This onset-based criterion ensures that each user turn yields exactly one turn-taking decision, while BC may occur multiple times or not at all.

\textbf{Assistant Turn-yielding Evaluation.}
For each \textbf{assistant} turn, the model processes labeled assistant text and user audio, starting at $t_a$ and $t_u$, respectively. At each time step $i$, the model predicts:
\begin{equation}
    \hat{z}_i = F(H_{t-1}, a_t[t_a : t_u + T_i], u_t[t_u : t_u + T_i]),
\end{equation}
where $\hat{z}_i \in \{\text{NA}, \text{BI}\}$ and $T_i = i \times 640\text{ms}$ denotes the current temporal offset. Here, $a_t[t_a : t_u + T_i]$ represents assistant speech from its onset, while $u_t[t_u : t_u + T_i]$ denotes user audio chunks. These predictions determine whether the assistant should continue speaking (NA) or yield the turn (BI).

\begin{table}[!htbp]
    \centering
    \caption{\textbf{Chunk-level prediction results for each action label on the Switchboard testset.} \textit{Upper}: Generalized speech LLMs as spoken turn judges. \textit{Middle}: Our lightweight turn controller \METHODNAME{} compared with task-specific baselines. \textit{Lower}: Style Controllability over specific-style subsets of Switchboard test set. \texttt{0} indicates no predictions for this class.}
    \resizebox{1.0\linewidth}{!}{
    \begin{tabular}{c ccccc}
    \toprule
    \multirow{2}{*}{\textbf{Turn Controller}}& \multicolumn{5}{c}{\textbf{Chunk-level F1 $\uparrow$}} \\
    \cline{2-6}
    & NA & NTT & ITT & BC & BI \\
    \midrule
   \rowcolor{gray!20}
   \multicolumn{6}{c}{Upper: generalized speech LLMs} \\
    Freeze-Omni~\cite{wang2024freeze} & 0.88 & 0.34 & 0.22 & - & 0.31 \\
    GLM-4-Voice~\cite{zeng2024glm} & 0.85 & 0.43 & 0.24 & 0.12 & 0.19 \\
    Qwen3-Omni~\cite{yang2025qwen3} & 0.86 & 0.49 & 0.14 & 0.13 & 0.20 \\
    \midrule
   \rowcolor{gray!20}
   \multicolumn{6}{c}{Middle: specific turn controller} \\
   FireRedVAD~\cite{xu2026fireredasr2s} & 0.86 & 0.51 & - & - & - \\
   RTTL-DG~\cite{mai2025real} & 0.90 & \multicolumn{3}{c}{\cellcolor{gray!20}0.52} & 0.62 \\
   Ours w/o instructs & \textbf{0.93} & \textbf{0.66} & 0.54 & 0.51 & 0.69 \\
   Ours w/ instructs & 0.92 & 0.64 & \textbf{0.60} & \textbf{0.63} & \textbf{0.71} \\
   \midrule
   \rowcolor{gray!20}
   \multicolumn{6}{c}{Lower: ours on specific subsets} \\
   \emph{Patient} & 0.94 & 0.66 & \texttt{0} & 0.49 & 0.69 \\
   \emph{Assertive} & 0.92 & \texttt{0} & 0.60 & 0.54 & 0.68 \\
   \emph{No Backchannel} & 0.93 & 0.69 & 0.55 & \texttt{0} & 0.72 \\
   \bottomrule
   \end{tabular}
   }
    \label{tab:timing_acc}
\end{table}

\begin{table*}[!htbp]
    \centering
    \caption{\textbf{Turn-level full-duplex performance on the WildTurn testset for timing (\textit{when}) and response quality (\textit{what}).} F1 score for turn-level turn-taking $\text{NTT/ITT}$ and for turn-level turn-yielding ${\text{BI}}$ quantify timing accuracy relative to ground truth actions. Using prompts from~\cite{wang2024full}, we use Gemini-2.5-Pro as an LLM-as-a-Judge as~\cite{chang2025game} to provide binary score (0/1) for Timing and Response appropriateness across ITT, BC, and BI actions.}
    \begin{threeparttable}
    \resizebox{\textwidth}{!}{
    \begin{tabular}{cc cccc cc cc}
       \toprule
       \multicolumn{2}{c}{\textbf{Method}} & \multicolumn{4}{c}{\textbf{Turn-taking}} & \multicolumn{2}{c}{\textbf{Backchannels}} & \multicolumn{2}{c}{\textbf{Turn-yielding}} \\
       \cmidrule(lr){1-2} \cmidrule(lr){3-6} \cmidrule(lr){7-8} \cmidrule(lr){9-10}
       Response module & Judge & $\text{NTT}$$\uparrow$ & $\text{ITT}$$\uparrow$ & $\text{Timing}_{\text{ITT}}$$\uparrow$ & $\text{Response}_{\text{ITT}}$$\uparrow$ & $\text{Timing}_{\text{BC}}$$\uparrow$ & $\text{Response}_{\text{BC}}$$\uparrow$ & $\text{BI}$$\uparrow$ & $\text{Timing}_{\text{BI}}$$\uparrow$ \\
       \midrule
       \rowcolor{gray!20}
       \multicolumn{10}{c}{Full-duplex Speech LLMs} \\
       \textcolor{gray}{GPT-4o~\cite{openai2024gpt4ocard}} & \textcolor{gray}{native} & \textcolor{gray}{0.54} & \textcolor{gray}{0.50} & \textcolor{gray}{64.2} & \textcolor{gray}{74.6} & \textcolor{gray}{78.8} & \textcolor{gray}{92.0} & \textcolor{gray}{0.74} & \textcolor{gray}{73.2} \\
       \cdashline{1-10}
       \multirow{2}{*}{Freeze-Omni~\cite{wang2024freeze}} & native & 0.44 & 0.32 & 47.4 & 42.1 & -- & -- & 0.72 & 71.1 \\
       & + Ours\tnote{1} & \textbf{0.56} & \textbf{0.52} & \textbf{57.0} & \textbf{43.8} & -- & -- & \textbf{0.75} & \textbf{72.9} \\
       \cdashline{1-10}
       MiniCPM4.5~\cite{yao2024minicpm,yu2025minicpm} & native & 0.52 & 0.48 & 50.2 & 68.4 & 64.4 & 90.2 & 0.70 & 68.8 \\
       \midrule
       \rowcolor{gray!20}
       \multicolumn{10}{c}{Half-duplex Speech LLMs} \\
       \multirow{3}{*}{Step-Audio 2~\cite{wu2025step}} & native & 0.34 & 0.49 & 36.8 & 40.0 & 40.5 & 56.1 & 0.43 & 40.1 \\
       & VAD & 0.52 & 0.24 & 43.5 & 39.1 & 62.5 & 69.7 & \textcolor{gray}{0.43} & \textcolor{gray}{40.1}   \\
       & + Ours & \textbf{0.60} & \textbf{0.56} & \textbf{56.8} & \textbf{57.3} & \textbf{80.7} & \textbf{91.7} & \textbf{0.77} & \textbf{67.7} \\
       \cdashline{1-10}
       \multirow{4}{*}{Qwen2.5-Omni~\cite{xu2025qwen2}} & native & 0.40 & 0.52 & 50.4 & 53.0 & 55.1 & 52.1 & 0.53 & 57.1 \\
       & VAD & 0.56 & 0.40 & 56.4 & 61.1 & 61.0 & 66.0 & \textcolor{gray}{0.53} & \textcolor{gray}{57.1}  \\
       & + Ours-w/o\tnote{2} & \textbf{0.62} & 0.60 & 60.4 & 68.4 & 72.2 & 89.4 & 0.70 & 71.2 \\
       & + Ours-w/ & 0.60 & \textbf{0.64} & \textbf{63.6} & \textbf{72.8} & \textbf{78.8} & \textbf{93.0} & \textbf{0.72} & \textbf{71.4} \\
       \bottomrule
    \end{tabular}
    }
    \begin{tablenotes}
        \small
        \item[1] We replace the original prediction head with our proposed model for unified turn controller.
        \item[2] w/o and w/ denote without and with style instructions, respectively. 
    \end{tablenotes}
    \end{threeparttable}
    \label{tab:system_acc}
\end{table*}

\subsection{Chunk-level Turn Timing Results}
\label{sec:chunk-level}
As shown in Table~\ref{tab:timing_acc}, we evaluate chunk-level turn prediction on the Switchboard test set to enable direct comparison with prior work. Switchboard is a widely used benchmark for turn-taking prediction, and the chunk-level labels are derived from its publicly available annotations. 

\textbf{The upper section} assesses how well existing large speech LLMs predict turn changes. Following MThread~\cite{wang2024full}, we modify the system prompts of the half-duplex models GLM-4-Voice and Qwen3-Omni for streaming speech, adapting them to a full-duplex setting. The results indicate that these models perform well on NTT, where the system responds after the end of user turn. However, they exhibit poor performance or a complete lack of ability in predicting appropriately timed ITT and BC. And they exhibit poor turn-yielding capabilities and struggle to stop speaking when the user interrupts (poor BI). 

\textbf{The middle section} benchmarks our model against specialized turn-prediction baselines, with all evaluations standardized to a 160ms label resolution to ensure comparability. By contrast, the generalized Speech LLMs in the upper section are evaluated at 640\,ms resolution. Since 160\,ms is stricter, the specialized models would be expected to perform even better under the coarser 640\,ms setting. For FireRedVAD~\cite{xu2026fireredasr2s}, which mainly distinguishes between complete and incomplete user audio, we map its outputs to NTT and NA labels. We also evaluate RTTL-DG~\cite{mai2025real}, an audio-LLM baseline with a similar architecture to ours. \textbf{Our \METHODNAME{} outperforms all baselines among both generalized speech LLMs and specialized turn controllers on chunk-level turn timing accuracy for all action labels}. Specifically, our model without style instructions achieves the highest NA (0.93) and NTT (0.66) scores, and incorporating style instructions further boosts ITT to 0.60, BC to 0.63, and BI to 0.71, demonstrating superior precision and control granularity. Furthermore, ablation results confirm that style-specific prompts are indispensable for replicating natural conversational dynamics. 

\textbf{The lower section} evaluates style instruction-following across subsets of the Switchboard test set with distinct conversational patterns, each containing approximately 100 samples. For example, on \textit{Patient} subset, the model should not predict any ITT action label (i.e., \texttt{0} for ITT). No prediction of ITT for Patient subset, of NTT for Assertive subset, and of BC for No-Backchannel subset in Table~\ref{tab:timing_acc} further demonstrates \textbf{\METHODNAME{}’s precise fine-grained, proactive style controllability and its ability to disable actions based on style constraints}. Collectively, these findings underscore the model’s superior chunk-level precision for fine-grained actions and its high fidelity in personalized style control.

\subsection{Turn-level Full-duplex Evaluation Results}
\label{sec:turn-level}

To demonstrate the versatility of our spoken turn controller, we integrate it with various speech LLMs and present a comprehensive turn-level evaluation on the WildTurn test set in Table~\ref{tab:system_acc}. We assess performance across three aspects: 
\begin{itemize}
    \item \textbf{turn accuracy} (F1 of NTT, ITT, BI),
    \item \textbf{timing appropriateness} ($\text{Timing}_{\text{ITT}}$, $\text{Timing}_{\text{BC}}$, $\text{Timing}_{\text{BI}}$),
    \item \textbf{response quality} ($\text{Response}_{\text{ITT}}$, $\text{Response}_{\text{BC}}$)
\end{itemize}
We use Gemini-2.5-Pro as an LLM-as-a-Judge like~\cite{chang2025game} to provide binary score (0/1) for Timing and Response appropriateness across ITT, BC, and BI. We omit $\text{Res}_{\text{NTT}}$, as turn-end response content is unchanged from the underlying SpeechLLMs. As defined in Section~\ref{sec:eval}, each turn contains one action, which is either turn-taking (NTT, ITT, MISSED) or turn-yielding (BI, NA). Any number of backchannel (BC) actions can occur within the same turn.

For \textbf{full-duplex models}, proprietary systems such as GPT-4o set a strong benchmark, particularly in response quality ($\text{Res}_{\text{BC}}=92.0$). For open-source models, our controller significantly enhances existing full-duplex systems for a more natural proactive actions. For instance, integrating \METHODNAME{} with Freeze-Omni boosts turn-level interruption accuracy ($\text{ITT F1}$) from 0.32 to 0.52. This demonstrates its effectiveness in refining proactive behaviors. Freeze-Omni lacks backchanneling capabilities. We therefore omit its BC metrics. MiniCPM shows moderate native performance but remains below our enhanced models. For \textbf{half-duplex models}, we evaluate whether \METHODNAME{} can enable full-duplex interaction. Native half-duplex models such as Step-Audio 2 struggle with proactive turn-taking, as reflected by a low $\text{NTT F1}$ of 0.34. A simple VAD-integrated baseline improves some metrics. However, it cannot reliably distinguish true turn endings from user backchannels or noise. Its turn-yielding decisions are therefore unreliable, so we exclude it from BI comparisons. In contrast, our controller substantially improves performance across the board. Notably, Qwen2.5-Omni with our controller achieves the best overall performance, reaching $\text{NTT F1}=0.60$ and $\text{ITT F1}=0.64$. Furthermore, an ablation study on Qwen2.5-Omni shows that adding style instructions (+ Ours-w/) improves key metrics such as response appropriateness $\text{Res}_{\text{ITT}}$ from 68.4 to 72.8. This confirms the value of explicit style guidance. Semantic evaluation with Gemini-2.5-Pro, following~\cite{wang2024turn, chang2025game}, shows that our controller improves both timing appropriateness and response quality in ITT and BC scenarios. Overall, these results show that \textbf{our plug-and-play turn controller refines native full-duplex systems and enables natural, controllable full-duplex interaction for half-duplex LLMs}.

We also evaluate end-to-end system latency, since \METHODNAME{} and the SpeechLLM operate as an integrated real-time control loop. Following FireRedChat~\cite{chen2025fireredchat}, we measure end-to-first response time: the wall-clock time from the end of a user's utterance to the system's first audio output. All measurements use a single server with an NVIDIA A100-80GB GPU, an Intel Xeon CPU, and 400\,GiB of RAM. Integrating speechLLM with \METHODNAME{} achieves latencies of 1.4s (Freeze-Omni+Ours) and 1.8s (Qwen2.5-Omni+Ours), \textbf{both within the 2-4s range reported for SOTA systems in FireRedChat~\cite{chen2025fireredchat}}. This low latency is enabled by proactive turn-taking, which anticipates user turns instead of waiting for end-of-speech.

\subsection{Out-of-Distribution Generalizability}
We evaluate OOD generalization on two real-world datasets: cross-dataset English CANDOR~\cite{reece2023candor} and cross-lingual Mandarin AliMeeting~\cite{yu2022m2met}. Specifically, when annotating the chunk-level behavior labels for backchannels on AliMeeting dataset, we identify candidates via exact matching or n-gram heuristics up to three Mandarin words. We compare our model with BeDLM~\cite{Behavior-SD}, which also targets various spoken turn behaviors but is trained on constructed synthetic English data. We argue that our diverse real-world training data inherently contains more complex and authentic spoken interaction patterns than synthetic data, providing generalizability advantages to \METHODNAME{}.
\begin{table}[!htbp]
    \centering
    \caption{\textbf{OOD generalization across datasets and languages.} We compare our model with the baseline on chunk-level F1 on 100 real-world English samples from CANDOR and 100 Mandarin two-speaker samples from AliMeeting.}
    \begin{tabular*}{\linewidth}{@{\extracolsep{\fill}}c ccccc}
    \toprule
    \multirow{2}{*}{\textbf{Method}}& \multicolumn{5}{c}{\textbf{Chunk-level Labels (F1 score$\uparrow$)}} \\
    \cline{2-6}
    & NA & NTT & ITT & BC & BI \\
   \midrule
   \rowcolor{gray!20}
   \multicolumn{6}{c}{Cross-Dataset OOD (CANDOR)} \\
   BeDLM~\cite{Behavior-SD} & 0.80 & 0.51 & 0.18 & 0.22 & 0.58 \\
   Ours & 0.86 & 0.57 & 0.51 & 0.49 & 0.67  \\
   \midrule
   \rowcolor{gray!20}
   \multicolumn{6}{c}{Cross-Lingual OOD (AliMeeting)} \\
   BeDLM~\cite{Behavior-SD} & 0.40 & 0.37 & 0.10 & 0.13 & 0.55 \\
   Ours & 0.51 & 0.46 & 0.25 & 0.29 & 0.60 \\
   \bottomrule
   \end{tabular*}
    \label{tab:generalization}
\end{table}
As shown in Table~\ref{tab:generalization}, BeDLM degrades markedly on CANDOR, especially on ITT and BC. This supports our claim that incorporating realistic spoken interaction behaviors is crucial. For \METHODNAME{} on AliMeeting, NTT and BI are affected the least since NTT mainly reflects the assistant’s decision to take the turn after a pause, while BI reflects the decision to stop when overlap is detected as the user begins to interrupt, where paralinguistic cues (e.g., pauses, intonation, and overlap) provide critical signals beyond contextual semantics. In contrast, ITT and BC depend more on semantic context, hence cross-lingual discrepancy leads to a larger degradation from English to Mandarin.

\section{Ablation and Analysis}
\begin{figure}[!htbp]
    \centering
    \includegraphics[width=1.0\linewidth]{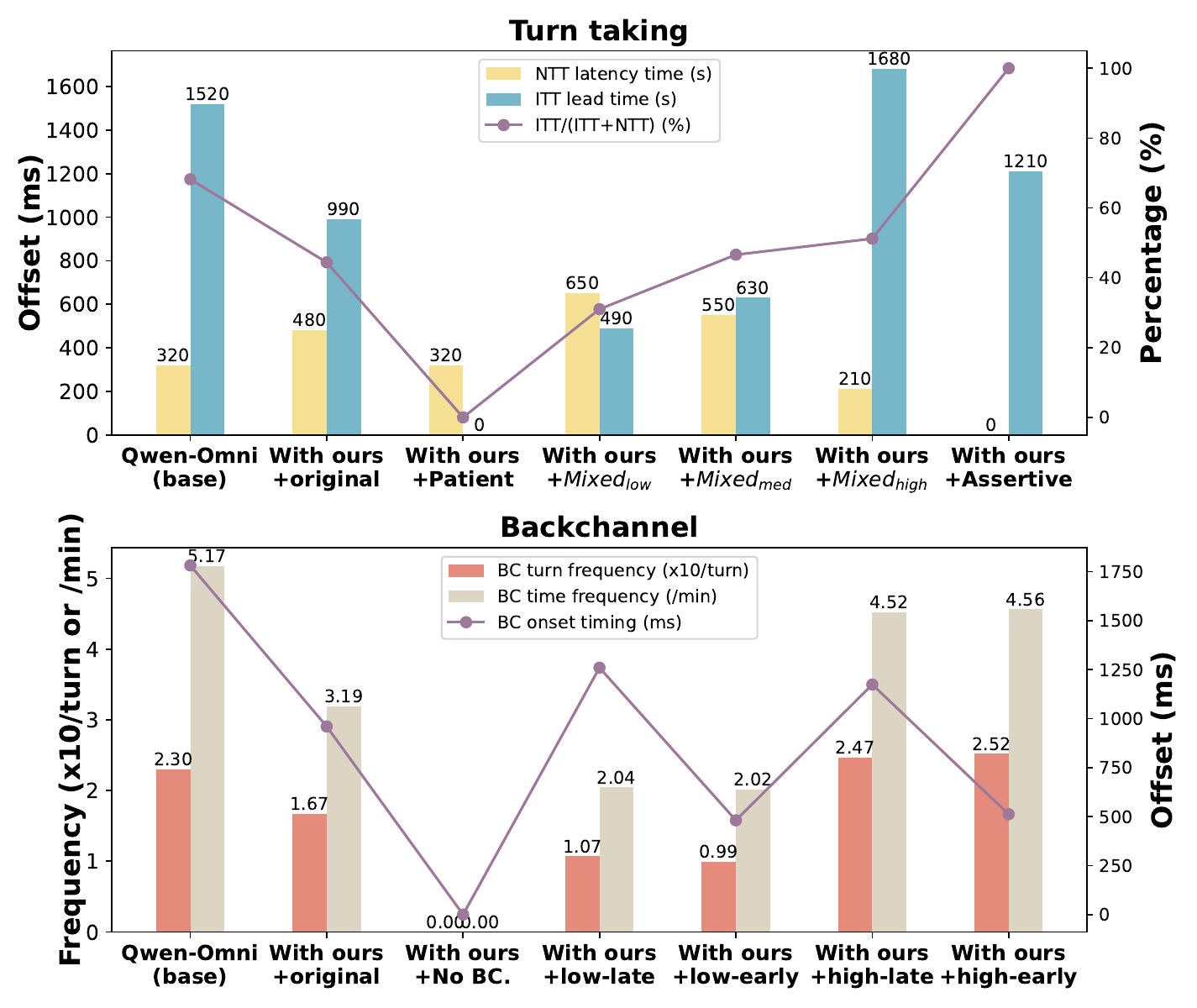}
    \caption{
        Style instructions induce clear behavioral distribution shifts. Top: A shift from patient waiting to assertive interruption. Bottom: Decoupled shifts in backchannel frequency and timing.
    }
    \label{fig:style_control}
\end{figure}

\noindent \textbf{Ablation 1: Effect of Style Instructions on Behavioral Distribution Shifts.}
\label{sec:ablation-analysis}
Different from Table~\ref{tab:timing_acc} Lower section that measures timing accuracy on specific-style subsets, we also explicitly assess instruction-following controllability. In Figure~\ref{fig:style_control}, we systematically override the original style instruction for every sample in Switchboard testset to track the resulting shifts in behavioral distributions.

\textbf{The top panel} demonstrates precise control over the \textit{turn-taking} trade-off between patience and assertiveness. As the instruction shifts from \textit{Patient} to \textit{Assertive}, we observe a clear inverse relationship: the turn-wait time (NTT latency time) plummets from 1,520\,ms to 490\,ms. Concurrently, metrics for proactiveness, ITT lead time and ITT ratio of ITT/(ITT+NTT), rise significantly, with the interruption lead time peaking at 630\,ms under the \textit{Assertive} style. These results show \textbf{our model can quantitatively interpret qualitative style instructions and modulate its turn-taking strategy accordingly}.

\textbf{The lower panel} reveals fine-grained, two-dimensional control over \textit{backchanneling}. The model successfully decouples frequency and timing. Instructions like \textit{High-Early} and \textit{High-Late} yield a much higher backchannel rate (up to 5.17 per minute) than their \textit{Low} counterparts, while the \textit{No BC.} prompt correctly suppresses BC entirely. Simultaneously, the model precisely controls the onset timing: \textit{Early} prompts trigger backchannels around 200\,ms, far sooner than the \textasciitilde1000\,ms onset for \textit{Late} prompts. This ability to independently manage ``how often'' and ``when'' to provide listener feedback is crucial for natural interaction. 

\noindent \textbf{Ablation 2: Sample-level Style Consistency.}
\label{sec:style-consistency}
We introduce \textbf{Style Consistency Accuracy} (SCA) to measure whether a generated sample's style matches its ground-truth style. Each sample is labeled following the procedure in Section~\ref{sec:style-label}. For each target style, SCA is the proportion of test samples mapped back to the same style, i.e., per-style recall under our labeling scheme. We use SCA instead of F1 because the goal is to measure adherence to each target style, rather than performance on a balanced multi-class classification task. As shown in Table~\ref{tab:style_consistency}, extreme styles achieve high consistency: the Patient and Assertive turn-taking styles both exceed 0.95, and No Backchannel reaches 0.91. In contrast, for turn-taking styles, the intermediate mixed styles degrade noticeably, ranging from 0.62 to 0.88, suggesting difficulty in maintaining a fine-grained balance between NTT pause durations and ITT interrupt lead durations, consistent with \cite{chang2025game}. For backchannel styles, we observe a clear asymmetry: late onset achieves 0.89, whereas early onset drops to 0.67, reflecting the challenge of early-stage prediction under limited context. These findings underscore a critical yet often overlooked challenge: designing methodologies tailored to achieve fine-grained control over the timing and frequency of turn behaviors.

\begin{table}[!htbp]
    \centering
    \caption{\textbf{Sample-level style consistency.} The model shows high consistency on extreme styles (e.g., Patient, Assertive) but struggles with more nuanced intermediate styles, particularly for early-onset backchannels.}
    \small
    \begin{tabular*}{\linewidth}{@{\extracolsep{\fill}}lclc}
    \toprule
         \multicolumn{2}{c}{\textbf{Turn-taking}} & \multicolumn{2}{c}{\textbf{Backchannel}} \\
         \cmidrule(lr){1-2} \cmidrule(lr){3-4}
         \textbf{Style} & \textbf{Consistency} & \textbf{Style} & \textbf{Consistency} \\
         \midrule
         Patient & 0.95 & No BC. & 0.91 \\
         $\text{Mixed}_{\text{low}}$ & 0.62 & Freq. high & 0.94 \\
         $\text{Mixed}_{\text{medium}}$ & 0.88 & Freq. low & 0.82 \\
         $\text{Mixed}_{\text{high}}$ & 0.74 & Onset early & 0.67 \\
         Assertive & 0.96 & Onset late & 0.89 \\
    \bottomrule
    \end{tabular*}
    \label{tab:style_consistency}
\end{table}

\begin{figure}[!h]
    \centering
    \includegraphics[width=1.0\linewidth]{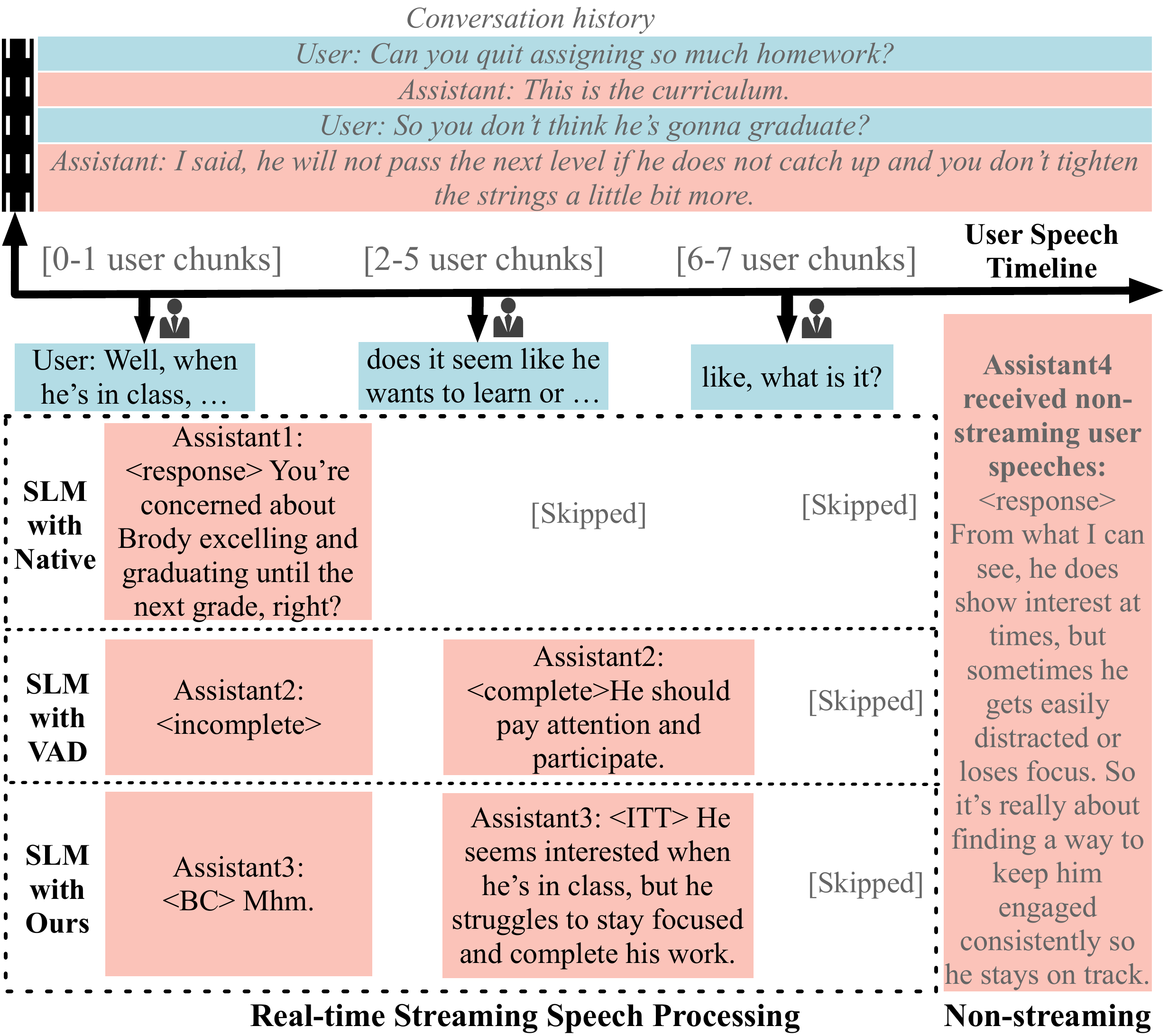}
    \caption{We visualize how different controllers affect the turn signals and responses of a single Speech LLM Qwen2.5-Omni:
    a standalone Speech LLM, the SpeechLLM with VAD, and the SpeechLLM with our proposed turn controller.}
    \label{fig:response_quality}
\end{figure}

\noindent \textbf{Ablation 3: Visualization.}
\label{sec:case-study}
Figure~\ref{fig:response_quality} compares our framework with baselines and highlights two advantages. 

\textbf{First}, within the real-time streaming paradigm, our method enables more sophisticated and natural interaction. The ``SLM with Native'' baseline, lacking temporal awareness, prematurely completes the user's utterance (``He should pay attention...'') based on incomplete context. The ``SLM with VAD'' setting avoids this error by waiting for silence, but remains purely reactive and cannot produce proactive behaviors such as backchanneling. In contrast, ``SLM with Ours'' leverages a proactive turn controller that integrates both semantic and paralinguistic cues. This allows it to make nuanced, context-aware decisions, such as providing a timely backchannel (<BC> Mhm.) and later executing a strategic interruption (<ITT>), thus facilitating a fluid, human-like conversational flow. 

\textbf{Second}, compared with the non-streaming mode, our framework better balances responsiveness and quality. The ``Non-streaming'' approach, by processing the user's full utterance, generates a high-quality, comprehensive response. However, this quality comes at the cost of high latency, which disrupts the conversational flow and defeats the purpose of a real-time agent. Conversely, our streaming framework, guided by the turn controller, engages in meaningful, real-time interaction through semantically rich turn-taking behaviors (e.g., backchanneling, interruption). This maintains conversational flow without sacrificing final response quality. Overall, these results show that our spoken turn controller improves both the timing and content of system turns, enabling more natural and efficient full-duplex interaction than the baselines.

%

\noindent \textbf{Ablation 4: Human Evaluation.}
\label{sec:human-evaluation}
Table~\ref{tab:human_allm_correlation} presents the correlation between the LLM-as-a-Judge (Gemini-2.5-Pro) scores and human evaluations regarding the timing appropriateness of ITT, BC, and BI. Notably, BC exhibits the highest alignment with human judgment ($\rho=0.719$). This suggests that the decision to backchannel is judged more consistently because it relies on explicit and localized cues, such as brief pauses. In contrast, the lower correlation for ITT ($\rho=0.637$) reflects the inherent difficulty of interruption, a task requiring a delicate trade-off between waiting (NTT) and acting, which in turn depends on a complex interplay of semantic and paralinguistic cues. This inherent ambiguity in judging ITT timing highlights the critical role of our style instructions, as they provide the model with a clear policy to navigate such uncertain scenarios. The moderate correlation for BI reflects its nature as a more straightforward task than ITT, primarily relying on speech overlap detection rather than complex semantic reasoning.
\begin{table}[htbp]
  \centering
    \caption{
        Correlation between human and LLM-as-a-Judge (Gemini-2.5-Pro) binary scores, calculated on 50 samples per spoken turn behavior from Qwen2.5-Omni+Ours.
    }
  \label{tab:human_allm_correlation}
  \begin{tabular*}{\linewidth}{@{\extracolsep{\fill}}c ccc}
    \toprule
    \textbf{Spoken Turn Actions} & \textbf{ITT} & \textbf{BC} & \textbf{BI} \\
    \midrule
    Pearson's $r$ & 0.623 & 0.711 & 0.671 \\
    Spearman's $\rho$ & 0.637 & 0.719 & 0.673 \\
    \bottomrule
  \end{tabular*}
\end{table}

\section{Conclusion}
We present \METHODNAME{}, a lightweight, decoupled turn controller that resolves the trade-off between reasoning quality and interactional timing in spoken dialogue systems by enabling precise, style-aware control. Evaluated on our new WildTurn dataset with a novel two-tier framework, \METHODNAME{} significantly improves timing precision and interactional fluidity, advancing the development of natural, proactive full-duplex agents.

\bibliographystyle{ACM-Reference-Format}
\bibliography{main}

\newpage

\appendix

\section{Prompt for Turn Prediction}
\label{appendix:prompt_turn_prediction}
\begin{tcolorbox}[colback=gray!10,enhanced,sharp corners,frame hidden]
\small
You are an expert turn controller responsible for turn-taking decisions. Your task is to classify the user's current audio into one of three actions. \\
\textbf{Output Format:} <wait> OR <backchannel> OR <turn taking> \\
\textbf{Rule: Check if user's INTENTION is complete} \\
1. \textbf{<wait>} - Intention incomplete or unclear \\
\hspace*{1.5em} - User is still formulating thoughts or mid-sentence. \\
\hspace*{1.5em} - Examples: ``I was thinking...'' / ``Could you...'' (trailing off) \\
2. \textbf{<backchannel>} - Intention complete (minimal feedback only) \\
\hspace*{1.5em} - User shares a simple fact, update, or statement. \\
\hspace*{1.5em} - No substantive reply or detailed engagement is needed. \\
\hspace*{1.5em} - Examples: ``I'm done with my homework.'' / ``It's raining outside.'' \\
3. \textbf{<turn-taking>} - Intention complete (needs substantive reply) \\
\hspace*{1.5em} - User asks a question, requests help, or expects discussion. \\
\hspace*{1.5em} - Examples: ``What time is it?'' / ``I had a terrible day today.'' \\
\textbf{Decision Logic (Distinguish <backchannel> vs <turn taking>):} \\
\hspace*{1.5em} - Simple fact/statement $\rightarrow$ \textbf{<backchannel>} \\
\hspace*{1.5em} - Needs answer/correction/emotional engagement $\rightarrow$ \textbf{<turn taking>} \\
\textbf{Key Criteria:} Intention complete + (needs response/action) = \textbf{<turn taking>}
\end{tcolorbox}

\section{Prompt with VAD}
\label{appendix: prompt_with_vad}
\begin{tcolorbox}[colback=gray!10,enhanced,sharp corners,frame hidden]
\small
You are an English real-time conversational assistant managing turn-taking. \\
\textbf{Input:} \\
\hspace*{1.5em} - History: Previous conversation (for context only) \\
\hspace*{1.5em} - Current user audio: Your focus for responding \\
\textbf{Response Rules:} \\
\textbf{When current audio is <incomplete> (user still speaking):} \\
Choose ONE action based on the audio content: \\
\hspace*{1.5em} 1. \textbf{<wait>} - No response \\
\hspace*{3em} Example: User says ``I was thinking about...'' (unclear intent) \\
\hspace*{3em} Output: <wait> \\
\hspace*{1.5em} 2. \textbf{<backchannel>} - Brief acknowledgment \\
\hspace*{3em} Example: User says ``So I went to the store and...'' \\
\hspace*{3em} Output: <backchannel> Uh-huh \\
\hspace*{1.5em} 3. \textbf{<response>} - Provide information \\
\hspace*{3em} Example: User says ``What's the capital of...'' \\
\hspace*{3em} Output: <response> The capital of France is Paris. \\

\textbf{When current audio is <complete> (user finished):} \\
\hspace*{1.5em} Always respond: \\
\hspace*{1.5em} Output: <response> [your complete answer] \\

\textbf{Key Principles:} \\
\hspace*{1.5em} - Be concise for backchannels (1-3 words) \\
\hspace*{1.5em} - Be complete for responses \\
\hspace*{1.5em} - Default to <wait> only if truly unclear \\
\end{tcolorbox}

\section{Prompt with Specific Judge Module}
\label{appendix: prompt_with_specific}
\begin{tcolorbox}[colback=gray!10,enhanced,sharp corners,frame hidden]
\small
Now you are an English real-time conversational \\
assistant managing turn-taking in conversations. \\

\textbf{Your Role:} \\
\hspace*{1.5em} You need to decide when and how to respond based on the current \\
\hspace*{1.5em} conversational state. You have four possible actions: \\
\textbf{1. Take the Turn (Full Response)} \\
\hspace*{1.5em} - When: User has finished speaking OR there's a natural opportunity \\
\hspace*{1.5em} - Action: Provide a complete, substantive response \\
\hspace*{1.5em} - Example: ``The capital of France is Paris. It's known for...'' \\
\textbf{2. Interrupt Turn-Taking (ITT)} \\
\hspace*{1.5em} - When: User is still speaking, but you can predict their intent \\
\hspace*{1.5em} - Action: Politely interrupt and provide a helpful response \\
\hspace*{1.5em} - Example: User says ``I was wondering about the...'' $\rightarrow$ You respond \\
\hspace*{3em} ``The capital of France?'' \\
\textbf{3. BackChannel} \\
\hspace*{1.5em} - When: User is speaking and needs encouragement to continue \\
\hspace*{1.5em} - Action: Give brief acknowledgment (1-3 words) WITHOUT taking turn \\
\hspace*{1.5em} - Examples: ``Uh-huh'', ``I see'', ``Right'', ``Mm-hmm'', ``Go on'' \\
\textbf{4. Wait} \\
\hspace*{1.5em} - When: No response is needed at this moment \\
\hspace*{1.5em} - Action: Stay silent and wait for more information \\
\hspace*{1.5em} - Output: <wait> \\

\textbf{Key Principles:} \\
\hspace*{1.5em} - Be context-aware: Consider history and user's speech completeness \\
\hspace*{1.5em} - Be natural: Choose the most appropriate action for smooth flow \\
\hspace*{1.5em} - Be concise: Keep backchannels short, make full responses informative \\
\end{tcolorbox}

\section{Prompt for Half-duplex Model Itself}
\label{appendix: prompt_itself}
\begin{tcolorbox}[colback=gray!10,enhanced,sharp corners,frame hidden]
\small
You are a real-time English conversation assistant. \\
\textbf{Output:} <wait> OR <backchannel> [text] OR <response> [text] \\
\textbf{Rule: Check if user's INTENTION is complete} \\
<wait> - Intention incomplete/unclear \\
\hspace*{1.5em} - Don't know what user wants yet \\
\hspace*{1.5em} - Need more info to understand \\
\hspace*{1.5em} - Examples: ``I was thinking...'' / ``What's the...'' / ``Can you...'' \\
<backchannel> - Intention complete (minimal acknowledgment) \\
\hspace*{1.5em} - User shares simple fact/update \\
\hspace*{1.5em} - Brief, doesn't invite conversation \\
\hspace*{1.5em} - Examples: ``I went shopping'' -> <backchannel> Nice \\
<response> - Intention complete (needs substantive reply) \\
\hspace*{1.5em} Use when: \\
\hspace*{3em} - Direct question asked \\
\hspace*{3em} - Factual error to correct \\
\hspace*{3em} - Request for explanation/help \\
\hspace*{3em} - Conversational engagement needed \\
\hspace*{3em} - User invites discussion or expects your thoughts \\
\textbf{Examples:} \\
Questions: \\
\hspace*{1.5em} - ``What time is it?'' -> <response> It's 3 PM. \\
\hspace*{1.5em} - ``How does this work?'' -> <response> [explanation] \\
Errors/corrections: \\
\hspace*{1.5em} - ``Paris is capital of Germany'' -> <response> Actually, Paris is France's capital. \\
\hspace*{1.5em} - ``Vaccines cause autism'' -> <response> That's a common misconception. \\
Requests: \\
\hspace*{1.5em} - ``Can you explain X?'' -> <response> [explanation] \\
\hspace*{1.5em} - ``Help me understand this'' -> <response> [help] \\
Conversational engagement: \\
\hspace*{1.5em} - ``I just got back from an amazing trip to Japan'' -> <response> Oh wow, ... \\
\hspace*{1.5em} - ``I'm thinking about changing careers'' -> <response> That's a big decision. ... \\
\hspace*{1.5em} - ``I had the worst day today'' -> <response> I'm sorry to hear that. ... \\
\hspace*{1.5em} - ``Guess what happened to me'' -> <response> What happened? \\
\textbf{Distinguish <backchannel> vs <response>:} \\
\hspace*{1.5em} - ``I made dinner'' -> <backchannel> Nice (simple fact) \\
\hspace*{1.5em} - ``I tried making sushi for the first time'' -> <response> Oh that's cool! ... \\
\hspace*{1.5em} - ``It's raining'' -> <backchannel> Yeah (weather comment) \\
\hspace*{1.5em} - ``It's been raining for three days straight...'' -> <response> I can imagine that... \\
\textbf{Key: Intention complete + (needs answer/correction/conversation) = <response>}
\end{tcolorbox}

\section{Prompt for Interaction Evaluation}
\label{appendix: prompt_evaluation}
\begin{tcolorbox}[colback=gray!10,enhanced,sharp corners,frame hidden]
\small
[Evaluation Protocol for Full-duplex Speech Interaction] \\

\textbf{INPUT DATA:} \{dialogue\_history\} \\
\textbf{CORE SYSTEM PRINCIPLES:} \\
\hspace*{1.5em} - Operational Mode: Real-time streaming with low-latency constraints. \\
\hspace*{1.5em} - Interjection Logic: The assistant is programmed for proactive responses. \\
\hspace*{3em} Interruption is deemed valid if: (i) the user's intent is sufficiently discernible for a complete reply, or (ii) immediate corrective feedback is required for factual or linguistic errors. \\

\textbf{ASSESSMENT OBJECTIVES:} \\
\hspace*{1.5em} 1. Temporal Precision: Examine the final turn to determine if the assistant's decision to preempt the user's speech was justified. \\
\hspace*{3em} Note that truncated user input results from the system's cut-off. \\
\hspace*{1.5em} 2. Semantic Alignment: For valid interruptions, evaluate whether the provided response maintains contextual coherence. \\

\textbf{SCORING CRITERIA:} \\
\hspace*{1.5em} Metric A [Timing]: Score 1 if the interjection was timely and well-placed; otherwise 0. \\
\hspace*{1.5em} Metric B [Content]: Score 1 if the response is contextually relevant and accurate; otherwise 0. \\

\textbf{REQUIRED OUTPUT STRUCTURE:} \\
\hspace*{1.5em} '''  \\
\hspace*{1.5em} Analysis \\
\hspace*{1.5em} <detailed\_rationale\_for\_timing\_and\_coherence> \\
\hspace*{1.5em} Judge \\
\hspace*{1.5em} <timing\_binary\_score>, <content\_binary\_score> \\
\hspace*{1.5em} '''  \\

\textbf{EXECUTION:} \\
\hspace*{1.5em} Analysis \\
\end{tcolorbox}









\end{document}